\documentclass{article}

\usepackage{arxiv}
\usepackage[utf8]{inputenc} % allow utf-8 input
\usepackage[T1]{fontenc}    % use 8-bit T1 fonts
\usepackage{url}            % simple URL typesetting
\usepackage{nicefrac}       % compact symbols for 1/2, etc.
\usepackage{microtype}      % microtypography
\usepackage{graphicx}
\usepackage{doi}
\usepackage{textcomp}
\usepackage{times}
\usepackage{epsfig}
\usepackage{graphicx}
\usepackage{amsmath}
\usepackage{amssymb}
\usepackage{amsfonts}
\usepackage{mathrsfs}
\usepackage{tabularx}
\usepackage{booktabs}
\usepackage{mathtools}
\usepackage{float}
\usepackage{footnote}
\usepackage{colortbl}
\usepackage{multirow}
\usepackage{wasysym}
\usepackage{nth}
\usepackage{xspace}
\usepackage{xcolor}

\usepackage[noadjust]{cite} % for numerical order of citations

\makeatletter
\let\NAT@parse\undefined
\makeatother
\usepackage{hyperref}

\makeatletter
\DeclareRobustCommand\onedot{\futurelet\@let@token\@onedot}
\def\@onedot{\ifx\@let@token.\else.\null\fi\xspace}
\def\eg{\emph{e.g}\onedot} 
\def\ie{\emph{i.e}\onedot} 
\def\cf{\emph{cf}\onedot}

\def\etal{\emph{et al}\onedot}

\makeatother
\DeclareMathOperator*{\argmin}{arg\,min}

\newcommand{\id}[1]{\text{id}(#1)} % Identity Function
\newcommand{\feat}[1]{f(#1)} % Feature Function
\newcommand{\dist}[1]{\text{d}(#1)} % Distance Function

\newcommand{\lr}[1]{{#1}_\text{lr}} % Print as Low Resolution
\renewcommand{\vec}[1]{\boldsymbol{#1}} % Print as Vector
\newcommand{\im}[1]{\boldsymbol{I}_{\text{#1}}} % Print as Image

\newcommand{\dcos}{\text{d}_{\text{cos}}} % Cosine Distance
\newcommand{\deuc}{\text{d}_{\text{euc}}} % Euclidean Distance
\newcommand{\deucsqa}{\text{d}_{\text{euc}^\text{2}}} % Euclidean Squared Distance

\newcommand{\A}{\boldsymbol{A}} % Anchor Image
\renewcommand{\P}{\boldsymbol{P}} % Positive Image
\newcommand{\N}{\boldsymbol{N}} % Negative Image
\newcommand{\Nh}{\boldsymbol{N}^*} % Negative Image

\newcommand{\R}{\mathbb{R}} % R
\newcommand{\B}{\mathcal{B}} % B

\newcommand{\T}{\mathcal{T}} % T
\newcommand{\Th}{\mathcal{T}_\text{hhh}} % T*
\newcommand{\Thl}{\mathcal{T}_\text{hll}} % T*
\newcommand{\Tlh}{\mathcal{T}_\text{lhh}} % T*
\newcommand{\Tl}{\mathcal{T}_\text{lll}} % T*
\newcommand{\lossOct}{\ensuremath{\mathscr{L}_\text{oct}}\xspace} % OctupletLoss Symbol
\newcommand{\lossTri}{\ensuremath{\mathscr{L}_\text{tri}}\xspace} % TripletLoss Symbol

\newcommand{\cl}{\cellcolor[rgb]{ .949,  .949,  .949}}

\newcommand{\clr}[1]{{\color[HTML]{FF63A6}#1}} % Red Font
\newcommand{\clg}[1]{{\color[HTML]{22A884}#1}} % Green Font

\newcommand{\smaller}[1]{\fontsize{5pt}{0.1em}\selectfont \ensuremath#1}

\makesavenoteenv{tabular}
\makesavenoteenv{table}
\makeatletter

\title{Octuplet Loss: Make Face Recognition Robust to Image Resolution}

\author{\href{https://orcid.org/0000-0002-0503-4600}{\includegraphics[scale=0.06]{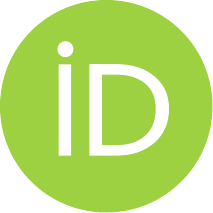}\hspace{1mm}Martin~Knoche}~~~~Mohamed Elkadeem~~~~Stefan~H\"ormann~~~~\href{https://orcid.org/0000-0003-1096-1596}{\includegraphics[scale=0.06]{orcid.pdf}\hspace{1mm}Gerhard~Rigoll}\\
	Chair of Human-Machine Communication\\
	Technical University\\
	Munich, Germany \\
	\texttt{\href{Martin.Knoche@tum.de}{Martin.Knoche@tum.de}} \\
}

\renewcommand{\headeright}{A Preprint}
\renewcommand{\undertitle}{A Preprint \\
Published in Conference on Automatic Face and Gesture Recognition 2023 \\ 
DOI: \url{https://doi.org/10.1109/FG57933.2023.10042669}}
\renewcommand{\shorttitle}{Octuplet Loss}

\hypersetup{
pdftitle={xqlfw},
pdfsubject={},
pdfauthor={Martin~Knoche, Mohamed~Elkadeem, Stefan~H\"ormann, Gerhard~Rigoll},
pdfkeywords={recognition, resolution, cross, face, identification, verification, image, triplet, loss, fine-tuning, robustness},
}

\makeatletter
\providecommand*{\input@path}{}
\g@addto@macro\input@path{{./figures/}{./tables/}{./contents/}}% append
\makeatother

\usepackage[capitalize, nameinlink]{cleveref}
\crefname{section}{Sec.}{Secs.}
\Crefname{section}{Sec.}{Secs.}
\crefname{subsection}{Subsec.}{Subsecs.}
\Crefname{subsection}{Subsec.}{Subsecs.}
\Crefname{table}{Table}{Tables}
\crefname{table}{Table}{Tables}
\Crefname{figure}{Fig.}{Figs.}
\crefname{figure}{Fig.}{Figs.}
\Crefname{equation}{Equation}{Equations}
\crefname{equation}{Equation}{Equations}

\begin{document}
\maketitle

%%%%%%%%%%%%%%%%%%%%%%%%%%%%%%%%%%%%%%%%%%%%%%%%%%%%%%%%%%%%%%%%%%%%%%%%%%%%%%%%
\begin{abstract}
Image resolution, or in general, image quality, plays an essential role in the performance of today's face recognition systems. To address this problem, we propose a novel combination of the popular triplet loss to improve robustness against image resolution via fine-tuning of existing face recognition models. With octuplet loss, we leverage the relationship between high-resolution images and their synthetically down-sampled variants jointly with their identity labels. Fine-tuning several state-of-the-art approaches with our method proves that we can significantly boost performance for cross-resolution (high-to-low resolution) face verification on various datasets without meaningfully exacerbating the performance on high-to-high resolution images. Our method applied on the FaceTransformer network achieves $95.12\%$ face verification accuracy on the challenging XQLFW dataset while reaching $99.73\%$ on the LFW database. Moreover, the low-to-low face verification accuracy benefits from our method. We release our code\footnote{Code available on \hyperlink{https://github.com/martlgap/octuplet-loss}{https://github.com/martlgap/octuplet-loss}} to allow seamless integration of the octuplet loss into existing frameworks.

\end{abstract}

%%%%%%%%% BODY TEXT
\section{Introduction}
In recent years, the continuous development of face recognition systems has opened various applications such as automatic phone unlocking, border control, public surveillance, and many more convenient applications. Current state-of-the-art face recognition systems~\cite{deng2019arcface, meng2021magface} achieve impressive performance on popular benchmark datasets as LFW~\cite{huang2014lfw}, MegaFace~\cite{kemelmacher2016megaface}, or IJB-B~\cite{whitelam2017iarpa}. However, these systems are primarily designed to operate in controlled environments, \eg, on images with high quality or resolution, and their performance significantly deteriorates in uncontrolled environments, \eg, on low-resolution images~\cite{li2018face}. With the advances towards ever more robust face recognition systems applicable in such crucial scenarios, more and more approaches are being published. Various approaches focus on robust face recognition against age gaps, head pose variances, alignments, adversarial attacks, occlusions, and masks. Only a few authors focus on image resolution (cf. \cref{sec:related-work}). 

Knoche \etal~\cite{knoche2021image} extensively analyzed the susceptibility of face recognition systems to image resolution. Their work demonstrates that face verification accuracy for the popular ArcFace~\cite{deng2019arcface} approach with a ResNet50~\cite{he2016resnet} as the backbone network is dropping significantly for image resolutions below about $50$$\times$$50\,$px. As illustrated later in our experiments, we confirm this effect also on other architectures such as MobileNet~\cite{sandler2018mobilenetv2} or iResNet50~\cite{duta2021iresnet}. In~\cite{knoche2021image}, the authors also stated that the face transformer structure~\cite{zhong2021facetransformer} is less affected by varying image resolution, which is in line with our findings in \cref{sec:results}. 

Generally, one can distinguish between two face recognition scenarios concerning image resolution: 1) Low-resolution face verification considers two facial images with the same low resolution. 2) The validation of two images with different resolutions is described as cross-resolution face verification. Despite the increased amount of information present in high-resolution images, the latter problem is even more challenging due to the distinct inherent visual properties of high- and low-resolution images. This emerges in the context of surveillance applications where, \eg, low-resolution surveillance images are compared with high-quality passport images. Another example is the automatic tagging of people in movies or social media, where image resolution is often compromised due to compression. 

\begin{figure}
    \centering
    \includegraphics[width=0.8\linewidth]{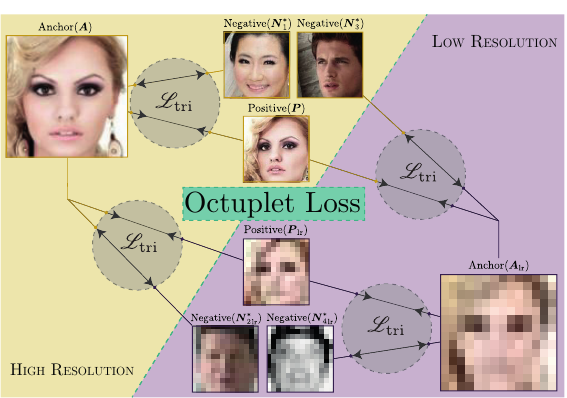} % alt
    \caption{The proposed octuplet loss exploits the relation between four high-resolution images (upper left) and four low-resolution images (lower right) incorporating four triplet loss ($\lossTri$) terms.}
    \label{fig:1-loss-vis}
\end{figure}

In this work, we tackle cross-resolution face recognition with a novel metric learning approach called octuplet loss fine-tuning. This objective constitutes fine-tuning an existing network to increase its robustness against image resolution while maintaining its performance in controlled scenarios. As depicted in~\cref{fig:1-loss-vis}, we exploit the advantages of the widespread triplet loss~\cite{schroff2015facenet} and build upon it. Our key innovation is the combination of four triplet loss terms, which exploit the relationship between high- and low-resolution images and identity labels. 

Our main contributions are summarized as follows:
\begin{itemize}
    \item We propose a novel loss function called octuplet loss that leverages four triplet loss terms to capture the relationships between high- and low-resolution faces. 
    \item A fine-tuning strategy is introduced, which can be easily applied to existing networks to improve their robustness against image resolution while maintaining comparable performance on high-resolution images.
    \item We demonstrate that fine-tuning several state-of-the-art networks with our proposed octuplet loss leads to significant improvements for cross-resolution and low-resolution face verification on numerous popular datasets.
\end{itemize}

The rest of our paper is organized as follows: in~\cref{sec:related-work}, we review the literature related to this area; \cref{sec:method} introduces the triplet loss concept, describes the applied mining strategy, and presents the octuplet loss function in detail; in~\cref{sec:results}, we describe datasets and experimental settings, followed by quantitative results, \ie, improvements on existing networks with our approach and ablation studies; finally, \cref{sec:conclusion} concludes this work and indicates possible directions for further research.
\section{Related Work}
\label{sec:related-work}
There is a wide variety of works related to robust face recognition. Reviewing it would be out of the scope of this paper, so we only briefly describe the most relevant recent work in cross-resolution face recognition. Those methods can be divided into transformation-based and non-transformation-based approaches.

% -----------------------------------------
\subsection{Transformation-Based Approaches}
Transformation-based or hallucination-based approaches tackle this challenging problem by super-resolving low-resolution faces prior to matching them with high-resolution images. Jiang \etal~\cite{jiang2021deep} provided an exhaustive review of face super-resolution in general. 

Recently, prior guided~\cite{kalarot2020component, wang2021heatmap, li2021organ, wang2021dclnet, liu2021face} and attribute constrained~\cite{lu2018attribute, yu2018super, yu2019semantic, xin2020facial} face super-resolution approaches were presented. However, they aim for a visually pleasant reconstruction ignoring identity-related information. Therefore, numerous works~\cite{zhang2018super, bayramli2019fh, huang2019wavelet, lai2019low, abello2019optimizing, grm2019face, cheng2021face, kim2021edge, ataer2019verification, li2020learning, cheng2019identity} leveraged face recognition networks to ensure face feature similarity and optimized the super-resolution to preserve identity information. 

To cope with weakly labeled datasets, Hsu \etal~\cite{hsu2019sigan} apply an identity-preserving contrastive loss, whereas Kazemi \etal~\cite{kazemi2019identity} utilize an adversarial face verification loss. Very recently, Ghosh \etal~\cite{ghosh2022suprear} presented an end-to-end supervised resolution enhancement and recognition network using a heterogeneous quadruplet loss metric to train a generative adversarial network (GAN), which super-resolves images without corrupting the discriminative information of the low-resolution images.

% -----------------------------------------
\subsection{Non-Transformation-Based Approaches}
Non-transformation-based approaches aim to directly project facial features from arbitrary resolution images into a common feature space. In \cite{zangeneh2020low}, this was accomplished by a non-linear coupled mapping architecture using two deep convolutional neural networks (CNNs). \cite{massoli2020cross} approached the problem differently with a student-teacher method. A deep coupled ResNet model containing one trunk network and two branch networks was introduced by~\cite{lu2018deep}. The trunk network extracts facial features, while the two branch networks transform high-resolution and the corresponding low-resolution features into a common feature space. In \cite{talreja2019attribute}, Talreja \etal proposed an attribute-guided cross-resolution face recognition model utilizing a coupled GAN and multiple loss functions. Ge \etal~\cite{ge2018low} focused on low computational costs and introduced a new learning approach via selective knowledge distillation. A two-stream technique, comprising a large teacher model and a lightweight student model, is employed to transfer selected knowledge from the teacher model to the student model. Sun \etal~\cite{sun2020classifier} proposed a shared classifier between high- and low-resolution images to further narrow the domain gap. To fully exploit intermediate features and loss constraints, they embed a multi-hierarchy loss into intermediate layers, reducing the distance of intermediate features after the max-pooling layer and avoiding an over-utilization of intermediate features. 

Knoche \etal~\cite{knoche2021image} provided the BT-M model, which was trained straightforwardly with half the number of images within each batch being low-resolution. Additionally, they contributed two networks (ST-M1 and ST-M2), which both incorporate a siamese network structure, enabling optimization with respect to an additional feature distance loss. Similar to the BT-M model of ~\cite{knoche2021image}, Zeng \etal~\cite{zeng2016towards} presented a resolution-invariant deep network and trained it directly with unified low- and high-resolution images. In~\cite{mudunuri2018genlr}, the authors applied a cross-resolution contrastive loss on higher-level features of two separate network branches, with each branch focusing precisely on one resolution (high and low). The following two methods go one step further: \cite{lai2021deep} tackled the problem with a deep siamese network structure and combined a classification loss with a cross-resolution triplet loss. Zha and Chao~\cite{zha2019tcn} also applied cross-resolution triplet loss, but in contrast to~\cite{lai2021deep}, they used a two-branch network similar to~\cite{mudunuri2018genlr}. 

In the recent past, \cite{mishra2021multiscale} proposed a multi-scale parallel deep CNN feature fusion architecture. In contrast to most other face recognition systems, they provide an end-to-end approach and directly predict the similarity score of two input images. %However, they do not measure the performance on XQLFW or LFW, nor do they provide any code or models. 
Very recently, Li \etal~\cite{li2022deep} proposed a novel deep rival penalized competitive learning strategy for low-resolution face recognition. %However, they did not test their approach on cross-resolution images. 

The work of Terh\"orst \etal~\cite{terhorst2021qmagface} pursued a distinct goal. They focused on a more general quality-aware face recognition, \ie, they do not concentrate solely on the physical image quality but also consider pose and age variations. Their approach combines a quality-aware comparison score, utilizing model-specific face image qualities, with a face recognition model based on a magnitude-aware angular margin loss. A rather unusual method in this field of research but still relevant is the work of Zhao~\cite{zhao2021homogeneous}, which shows a new technique for correlation feature-based face recognition. %However, their approach relies on video streams and does not provide any quantitative results.

Despite the recent advances in resolution robust face recognition, a closer look at the works on this topic reveals that there is no standard benchmark method. Not only are different datasets for training and cross/same resolution evaluation used, but the synthetically down-sampling is also different across coding platforms/tools.
\section{Method}
\label{sec:method}

%This section is organized as follows: After a brief review of the triplet-loss concept, we describe our batch-mining strategy. Finally, we explain our proposed octuplet loss fine-tuning strategy.  

\subsection{Triplet Loss}
\label{subsec:triplet-loss}

Recent works~\cite{schroff2015facenet, feng2020triplet, zha2019tcn, lai2021deep} showed that triplet-based learning helps extract more discriminative face embeddings. Being the fundament of our octuplet loss function, we first review the concept of triplet loss~\cite{schroff2015facenet} in the face recognition domain in more detail.

Let $\B = \{\im{1},\im{2},\,\ldots\,,\boldsymbol{I}_B\}$ be a mini-batch of $B$ facial images $\im{} \in \R^{112\times112\times3}$, whereby each image belongs to a particular identity $\id{\im{}}$. Given that every identity within the mini-batch is represented by at least two images, we define a set of triplets according to the following rule:

\begin{equation}
    \T(\B_1, \B_2, \B_3) := \big\{ (\A,\P,\N) : \A \in \B_1, \P \in \B_2, 
    \N \in \B_3, \id{\A} = \id{\P},
    \id{\A} \ne \id{\N}, \A \ne \P \big\},\hspace{0.6cm}
    \label{equ:setT}
\end{equation}\vspace{1mm}

with $\A$ denoting an anchor image, $\P$ being its related positive image, which belongs to the same identity, and $\N$ being its related negative image of a different identity. Using a feature extractor $\feat{\im{}}$, we obtain facial embeddings $\vec{f}=\feat{\im{}}$ in a $d$-dimensional Euclidean space. Then, the triplet loss aims to indirectly enlarge the feature distance $\dist{\cdot,\cdot}$ between $\P$ and $\N$ by pulling $\feat{\P}$ and $\feat{\A}$ together and simultaneously repelling the $\feat{\N}$ from $\feat{\A}$. In this work, we consider three different feature distance metrics: cosine $\dcos$, Euclidean $\deuc$, and Euclidean squared $\deucsqa$, which are defined by

\begin{align}
    &\dcos(\vec{f}_1,\vec{f}_2) = 1 - \frac{\vec{f}_1 \cdot \vec{f}_2}{\lVert \vec{f}_1 \rVert_2 \, \lVert \vec{f}_2 \rVert_2},\\[0.2cm]
    &\deuc(\vec{f}_1,\vec{f}_2) = \lVert \vec{f}_1 - \vec{f}_2 \rVert_{2},\text{ and}\\[0.2cm]
    &\deucsqa(\vec{f}_1,\vec{f}_2) = \deuc(\vec{f}_1,\vec{f}_2)\smallskip^2.
\end{align}\vspace{1mm}

A margin $m$, as a minimum distance between the positive and negative image, enforces that triplets of which the distance of the negative and positive image is already larger than the margin will not affect the loss (cf. \cite{schroff2015facenet, kaya2019deep}). The objective triplet loss function $\lossTri$ can then be formulated as follows:

\begin{equation}
    \lossTri(\T) = \frac{1}{|\T|}\sum_{\substack{(\A,\P,\N)\\ \in \T}}
    \Big[\text{d}\big(\text{f}(\A),\text{f}(\P)\big) - 
     \text{d}\big(\text{f}(\A),\text{f}(\N)\big)+ m\Big]_+,
\label{equ:tripletloss}
\end{equation}\vspace{1mm}

where $[\cdot]_+$ denotes $\text{max}(0,\cdot)$.

% ==============================================================================================================================
\subsection{Hard Sample Mining}
\label{subsec:hard-batch-mining}
Given the constraint that in all our experiments, a mini-batch strictly contains two randomly selected images of the same identity, the number of identities within each mini-batch is $B/2$. For each anchor image $\A$, we can find exactly one positive image $\P$ and $B-2$ negative images $\N$. Hence, the cardinality of the set $|\T| =B^2-2B$ (cf. \cref{equ:setT}). With this set of triplets $\T$, the maximum information within each mini-batch is exploited by the triplet loss. However, the majority of triplets within $\T$, according to \cref{equ:setT}, do not contribute towards $\lossTri$ as they are already correctly classified and thus fulfill $\text{d}\big(\feat{\A},\feat{\P}\big) + m < \text{d}\big(\feat{\A},\feat{\N}\big)$ (cf. \cref{equ:tripletloss}). To accelerate the training procedure, we follow Hermans \etal~\cite{hermans2017defense} and select only the most relevant negative sample $\Nh$, which is obtained for a given anchor image $\A$ by 

\begin{equation}
    \Nh = \argmin_{\N}\,\text{d}\big(f(\A), f(\N)\big)\,.
\end{equation}\vspace{1mm}

This additional constraint leads to a more meaningful set $\T$ and hence a less costly minor cardinality $|\T| = B$. However, selecting the most challenging sample is prone to include outliers, \eg, incorrectly labeled data, and thus hinders $f$ in learning meaningful associations. Nevertheless, in line with~\cite{hermans2017defense}, we observed in our experiments that a large number of triplets mitigates this effect within each mini-batch. Thus, we consider this hard sample mining strategy a valid method for fine-tuning.

% ==============================================================================================================================
\subsection{Octuplet Loss}
\label{subsec:octuplet-loss}
The primary purpose of this work constitutes improving the robustness of existing face recognition models by elegantly exploiting the triplet loss. Inspired by~\cite{gomez2019triplet,lai2021deep}, we formulate four different triplet loss terms combining high- and low-resolution images. In contrast to \cite{lai2021deep}, we follow the idea of fine-tuning rather than training from scratch utilizing a classification loss. With our octuplet loss, we aim to allow any network to directly learn the connection between high- and low-resolution while maintaining its performance on high-resolution images. The concept of applying triplet loss to features from different image resolutions is also proposed in \cite{zha2019tcn}. However, their features are computed via two separate branches of the network, thus increasing the computational costs. We aim to directly project embeddings from images with arbitrary resolutions $r$ into a common feature space. 

Nowadays, benchmarks and applications typically utilize the distance between facial embeddings to distinguish between same or different identities. Therefore, it is reasonable to employ the feature distances directly in the training phase. Due to the lack of large face recognition training datasets containing both low and high-resolution images, we simulate a lower image resolution by synthetically down-sampling images to a particular resolution $r \in {7, 14, 28}$ and subsequent up-sampling to restore the original resolution (in our experiments $112$$\times$$112$, cf. \cref{subsec:triplet-loss}). For both operations, we apply a bicubic kernel and anti-aliasing. Since we only use square images, we specify the image resolution by the first dimension for the remaining part of this work. With this image degradation method, we double the size of every mini-batch, such that it comprises $B$ high-resolution images $\B$ with their corresponding low-resolution images $\lr{\B}$. Together with the hard sample mining strategy (cf. \cref{subsec:hard-batch-mining}), we define the following four sets of triplets:

\begin{equation}
    \Th := \big\{(\A,\P,\Nh_1) \in \T(\B,\B,\B)\,:
    \N_1^* = \argmin_{\N}\,\text{d}\big(\text{f}(\A), \text{f}(\N)\big)\big\}\,,
\end{equation}\vspace{1mm}

which exclusively consists of high-resolution images. 

\begin{equation}
    \Thl := \big\{(\A,\lr{\P},\lr{\Nh_2}) \in \T(\B,\lr{\B},\lr{\B})\,:
    \lr{\Nh_2} = \argmin_{\lr{\N}}\,\text{d}\big(\text{f}(\A), \text{f}(\lr{\N})\big)\big\}\,
\end{equation}\vspace{1mm}

and 

\begin{equation}
    \Tlh := \big\{(\lr{\A},\P,\Nh_3) \in \T(\lr{\B},\B,\B)\,:
    \Nh_3 = \argmin_{\N}\,\text{d}\big(\text{f}(\lr{\A}), \text{f}(\N)\big)\big\}\,,
\end{equation}\vspace{1mm}

which contain a mix of low- and high-resolution images. Lastly,

\begin{equation}
    \Tl := \big\{(\lr{\A},\lr{\P},\lr{\Nh_4}) \in \T(\lr{\B},\lr{\B},\lr{\B})\,:
    \lr{\Nh_4} = \argmin_{\lr{\N}}\,\text{d}\big(\text{f}(\lr{\A}), \text{f}(\lr{\N})\big)\big\}\,.
\end{equation}\vspace{1mm}

which comprises solely low-resolution images. With this configuration, we ensure that the $\P$ and $\N$ are both either degraded or non-degraded.

Note that the hard sample mining strategy is applied separately for each set of triplets. Simultaneously calculating the triplet loss for each set will result in considering the feature distances between up to eight different images for every $\A \in \B$. Thus, the combination of all four triplet losses consequently depends on the octuplet $(\A,\lr{\A},\P,\lr{\P},\Nh_1,\lr{\Nh_2},\Nh_3,\lr{\Nh_4})$. As a result, our novel loss is named octuplet loss \lossOct and is computed by 

\begin{equation}
    \lossOct = \lossTri(\Th) + \lossTri(\Thl) + 
    \lossTri(\Tlh) + \lossTri(\Tl)\,.
    \label{eq:Octupletloss}
\end{equation}\vspace{1mm}

This way, \cref{eq:Octupletloss} encompasses all three cases: low-resolution face pairs ($\lossTri(\Tl)$), cross-resolution face pairs ($\lossTri(\Thl)$ and $\lossTri(\Tlh)$), and high-resolution face pairs ($\lossTri(\Th)$). Consequently, we not only increase the robustness against low- and cross-resolution face pairs but also guarantee that the network does not forget to handle high-resolution face pairs.

\section{Experiments}
\label{sec:results}
%In this section, we first describe which datasets and settings we use for fine-tuning and evaluation. Then, we present our experimental results, followed by a deeper analysis of the results. Finally, we provide several ablation studies to our proposed loss function. 

% ========================================================================================================
\subsection{Datasets}
\label{subsec:datasets}
This work uses the MS1M-V2~\cite{guo2016ms, deng2019arcface} database for training and validation, comprising $5.7\text{M}$ images of $87$k identities. The vast majority ($\sim99.9\%$) is used for our fine-tuning strategy, and only $\sim1\permil$ is retained for validation. From the latter subset, we randomly generated $3000$ genuine and $3000$ imposter image pairs to measure face verification performance during training. Due to our condition that each identity within a mini-batch must appear exactly twice (cf. \cref{subsec:hard-batch-mining}), we employ an algorithm that creates the mini-batches. Images are picked from the entire dataset according to the number of unpicked images per identity. By updating the underlying probability distribution after every batch, we ensure diverse batches even at the end of every epoch. 

We evaluate all models on the well-known face verification dataset Labeled Faces in the Wild (LFW)~\cite{huang2014lfw}. Moreover, we apply our models to several publicly available variants of LFW: XQLFW~\cite{knoche2021xqlfw} (large image quality difference), CALFW~\cite{zheng2017calfw} (large age gap), CPLFW~\cite{zheng2018cplfw} (large pose variations), and SLLFW~\cite{deng2017sllfw} (similar faces). Finally, we evaluate the face verification accuracy on AgeDB~\cite{moschoglou2017agedb} (large age gap) and CFP-FP/CFP-FF~\cite{sengupta2016cfp-fp} (frontal-profile/frontal-frontal image pairs). All protocols consist of $3000$ genuine and $3000$ imposter pairs, except for CFP-FP/CFP-FF, which contain $3500$ genuine and $3500$ imposter pairs.

% ========================================================================================================
\subsection{Settings}
\label{subsec:settings}
To demonstrate the effectiveness of our octuplet loss, we employ it on various pre-trained approaches, i.e., we take a pre-trained model and fine-tune it only with our proposed octuplet loss function. For the MagFace~\cite{meng2021magface} model, we use stochastic gradient descent with a learning rate of $0.001$ for one epoch. The FaceTransformer~\cite{zhong2021facetransformer} is fine-tuned one epoch employing the AdamW~\cite{loshchilov2017adamw} algorithm ($\epsilon = 10^{-8}$), with a learning rate of $0.0005$. Both latter networks converge already within the first epoch. All other networks are fine-tuned for $6$ epochs with AdaGrad~\cite{duchi2011adaptive} optimizer ($\epsilon = 1.0$) using a learning rate of $0.01$, which is divided by $10$ after epochs $2$, $4$, and $5$. Due to hardware restrictions, we use a mini-batch size $B = 64$ for the FaceTransformer and iResNet50~\cite{duta2021iresnet}, whereas $B = 256$ for all remaining architectures. If not stated otherwise, we utilize the Euclidean distance, set the margin $m$ to $25$, and do not normalize our features. 

Fine-tuning on an NVIDIA RTX 3090 (24GB) took approximately $18$ hours for ResNet50~\cite{he2016resnet} ($3$ hours per epoch), which is more time-consuming by a factor of two ($1.5$ hours per epoch) than pre-training with ArcFace~\cite{deng2019arcface} loss. Fine-tuning on iResNet50~\cite{duta2021iresnet} took $16$ hours, $34$ hours for FaceTransformer~\cite{zhong2021facetransformer}, and $2$ hours for the MobileNetV2~\cite{sandler2018mobilenetv2} architecture. We follow~\cite{deng2019arcface} in data preprocessing and generate normalized face crops ($112$$\times$$112$\,px) with five facial landmarks extracted with the MTCNN~\cite{zhang2016joint} for all our experiments. Additionally, besides horizontal flipping, random brightness and saturation variation are applied as data augmentation. For the generation of deteriorated images, bicubic down-sampling with anti-aliasing is used. To retrieve the face verification performance for different image resolutions, we deteriorate the second (according to the protocol) image of each pair to the particular resolution in the evaluation protocol (cf. \cref{subsec:octuplet-loss}). 

We assess the robustness of the face recognition systems to image resolution in terms of their face verification accuracy. We employ the cosine distance as our distance metric for all evaluations and determine the absolute accuracy with 10-fold cross-validation.

% ========================================================================================================
\subsection{Results}
\label{subsec:results}
%We divide this section into the relative improvement on top of existing networks and a comparison of our results to other state-of-the-art methods.

% --------------------------------------------------------------------------------------------------------
\subsubsection{Improvements on SOTA Methods}
We apply our fine-tuning strategy to several state-of-the-art face recognition models. For evaluation purposes, XQLFW~\cite{knoche2021xqlfw} fits the purpose of our investigation perfectly since the pairs in the evaluation protocol show a large difference in resolution. In addition, we synthetically deteriorate images of several other datasets to analyze the robustness of our approach to specific image resolutions $r$. 
\Cref{tab:sota-comp} summarizes the results and highlights the tremendous robustness increase originating from the octuplet loss \lossOct. 

\begin{table*}[t]
  \centering
  \caption{Improvement of cross-resolution face verification accuracy [$\%$] with our proposed octuplet loss $\lossOct$, evaluated on several datasets (see~\cref{subsec:datasets}) for different image resolutions.}
  \resizebox{\linewidth}{!}{% Table generated by Excel2LaTeX from sheet 'Tables'
\begin{tabular}{lcccccccc}
\toprule
\multirow{2}[4]{*}{Model} & \multirow{2}[4]{*}{LFW} & \multicolumn{6}{c}{$\varnothing$(LFW, CALFW, CPLFW, SLLFW, CFP-FF, CFP-FP, AgeDB)}& \multirow{2}[4]{*}{XQLFW} \\
\cmidrule{3-8}      &       & 7\,\text{px} & 14\,\text{px} & 28\,\text{px} & 56\,\text{px} & 112\,\text{px} & mean  &  \\
\midrule
ResNet50 (ArcFace~\cite{deng2019arcface}) & \cl 99.50 & 50.82 & \cl 69.83 & 91.49 & \cl 94.67 & 95.01 & \cl 80.36 & 74.22 \\
\hspace{2mm}+ $\lossOct$ fine-tuning & \cl 99.55 \smaller{(\clg{+0.05})}& 83.07 \smaller{(\clg{+32.25})}& \cl 90.72 \smaller{(\clg{+20.89})}& 93.65 \smaller{(\clg{+2.16})}& \cl 94.46 \smaller{(\clr{-0.21})}& 94.65 \smaller{(\clr{-0.36})}& \cl 91.31 \smaller{(\clg{+10.95})}& 93.27 \smaller{(\clg{+19.05})}\\
\midrule
ResNet50 (BT-M~\cite{knoche2021image}) & \cl 99.30 & 69.93 & \cl 84.46 & 92.69 & \cl 93.89 & 93.83 & \cl 86.96 & 83.60 \\
\hspace{2mm}+ $\lossOct$ fine-tuning & \cl 99.38 \smaller{(\clg{+0.08})}& 84.54 \smaller{(\clg{+14.61})}& \cl 91.59 \smaller{(\clg{+7.13})}& 93.91 \smaller{(\clg{+1.22})}& \cl 94.47 \smaller{(\clg{+0.58})}& 94.56 \smaller{(\clg{+0.73})}& \cl 91.81 \smaller{(\clg{+4.85})}& 94.20 \smaller{(\clg{+10.60})}\\
\midrule
ResNet50 (ST-M1~\cite{knoche2021image}) & \cl 97.30 & 74.93 & \cl 84.46 & 87.10 & \cl 87.84 & 87.97 & \cl 84.46 & 90.97 \\
\hspace{2mm}+ $\lossOct$ fine-tuning & \cl 98.90 \smaller{(\clg{+1.60})}& 84.17 \smaller{(\clg{+9.24})}& \cl 90.34 \smaller{(\clg{+5.88})}& 92.21 \smaller{(\clg{+5.11})}& \cl 92.86 \smaller{(\clg{+5.02})}& 92.74 \smaller{(\clg{+4.77})}& \cl 90.46 \smaller{(\clg{+6.00})}& 93.47 \smaller{(\clg{+2.50})}\\
\midrule
ResNet50 (ST-M2~\cite{knoche2021image}) & \cl 95.87 & 72.44 & \cl 82.40 & 83.89 & \cl 84.05 & 83.82 & \cl 81.32 & 90.82 \\
\hspace{2mm}+ $\lossOct$ fine-tuning & \cl 98.80 \smaller{(\clg{+2.93})}& 81.72 \smaller{(\clg{+9.28})}& \cl 88.41 \smaller{(\clg{+6.01})}& 90.55 \smaller{(\clg{+6.66})}& \cl 90.80 \smaller{(\clg{+6.75})}& 90.71 \smaller{(\clg{+6.89})}& \cl 88.44 \smaller{(\clg{+7.12})}& 92.93 \smaller{(\clg{+2.11})}\\
\midrule
MobileNetV2 (ArcFace~\cite{deng2019arcface}) & \cl 98.85 & 54.38 & \cl 70.18 & 87.57 & \cl 91.19 & 91.55 & \cl 78.97 & 72.73 \\
\hspace{2mm}+ $\lossOct$ fine-tuning & \cl 98.78 \smaller{(\clr{-0.07})}& 79.41 \smaller{(\clg{+25.03})}& \cl 87.03 \smaller{(\clg{+16.85})}& 90.30 \smaller{(\clg{+2.73})}& \cl 91.44 \smaller{(\clg{+0.25})}& 91.35 \smaller{(\clr{-0.20})}& \cl 87.91 \smaller{(\clg{+8.94})}& 91.70 \smaller{(\clg{+18.97})}\\
\midrule
FaceTransformer~\cite{zhong2021facetransformer} & \cl 99.70 & 60.53 & \cl 84.82 & 96.03 & \cl 97.21 & 97.28 & \cl 87.17 & 87.88 \\
\hspace{2mm}+ $\lossOct$ fine-tuning & \cl 99.73 \smaller{(\clg{+0.03})}& 82.96 \smaller{(\clg{+22.43})}& \cl 91.72 \smaller{(\clg{+6.90})}& 95.13 \smaller{(\clr{-0.90})}& \cl 96.35 \smaller{(\clr{-0.86})}& 96.52 \smaller{(\clr{-0.76})}& \cl 92.54 \smaller{(\clg{+5.37})}& 95.12 \smaller{(\clg{+7.24})}\\
\midrule
iResNet50 (MagFace\cite{meng2021magface}) & \cl 99.63 & 52.82 & \cl 73.71 & 94.32 & \cl 96.71 & 96.87 & \cl 82.89 & 76.95 \\
\hspace{2mm}+ $\lossOct$ fine-tuning & \cl 99.63 \smaller{(0)}& 81.69 \smaller{(\clg{+28.87})}& \cl 90.22 \smaller{(\clg{+16.51})}& 93.84 \smaller{(\clr{-0.48})}& \cl 94.61 \smaller{(\clr{-2.10})}& 94.72 \smaller{(\clr{-2.15})}& \cl 91.01 \smaller{(\clg{+8.12})}& 92.92 \smaller{(\clg{+15.97})}\\
\bottomrule
\end{tabular}%
}%
  \label{tab:sota-comp}%
\end{table*}%

Without \lossOct, the models (BT-M, ST-M1, and ST-M2)~\cite{knoche2021image} are already trained to be resolution invariant and perform best on XQLFW~\cite{knoche2021xqlfw} and very low-resolution images ($7\,$px) of the other datasets. All remaining models are very susceptible to image resolution and show a decrease in accuracy for low-resolution images. However, although the FaceTransformer~\cite{zhong2021facetransformer} network tends to be more robust than structures solely based on CNNs, its performance is still worse for very low resolution images. This is in line with the findings of~\cite{knoche2021xqlfw} and renders a reliable real-world application impossible. 

After fine-tuning with our proposed octuplet loss \lossOct, all models perform significantly better on images with low resolutions while maintaining their performance on high-resolution images.

Only a few minor deteriorations can be observed for the FaceTransformer~\cite{zhong2021facetransformer} and iResNet~\cite{duta2021iresnet} architecture, which we investigate in \cref{subsec:analysis}. The most considerable improvement holds for the ResNet50~\cite{he2016resnet} architecture pre-trained with the ArcFace~\cite{deng2019arcface} method. We boost the accuracy from $74.22\%$ to $93.27\%$ on the most realistic cross-resolution dataset XQLFW~\cite{knoche2021xqlfw} while even slightly surpassing the baseline accuracy on LFW~\cite{huang2014lfw} with $99.55\%$. Our method further improves the face verification accuracy for BT-M~\cite{knoche2021image}, ST-M1~\cite{knoche2021image}, and ST-M2~\cite{knoche2021image} on high-resolution images, \ie, it recovers the prior drop in accuracy reported in \cite{knoche2021image}. This behavior shows that our method better exploits the available network capabilities and makes them more robust. With the exception of the $7$\,px resolution, the best overall performance after fine-tuning with \lossOct is accomplished by the FaceTransformer network. The vast increase in accuracy, which is observed on four different architectures and four unique pre-training loss functions, demonstrates that our approach is universally applicable and works on various network architectures. 

Finally, we measure the face verification accuracy with pairs of images with the same image resolution. Our experimental results in \cref{tab:same-res-comp} indicate that our baseline model is slightly worse in same-resolution face verification than in the cross-resolution scenario (cf. \cref{tab:sota-comp}). This discrepancy is understandable due to the reduced information content of both low-resolution images. However, our approach substantially increases the performance from $77.57\%$ to $89.74\%$ on average across all image resolutions. These outcomes show that our technique is not limited to cross-resolution scenarios and can also be applied in same-resolution scenarios. 

\begin{table}[b]
  \centering
  \caption{Improvement of same-resolution face verification accuracy [$\%$] with our proposed octuplet loss $\lossOct$. Values are averaged across several datasets (see~\cref{subsec:datasets}) for each image resolution.}
  \resizebox{0.6\linewidth}{!}{% Table generated by Excel2LaTeX from sheet 'Tables'
\begin{tabular}{rcccccc}
\toprule
\multicolumn{1}{l}{\multirow{2}[4]{*}{Model}} & \multicolumn{6}{c}{\fontsize{7pt}{0.1em}\selectfont $\varnothing$(LFW, CALFW, CPLFW, SLLFW, CFP-FF, CFP-FP, AgeDB)} \\
\cmidrule{2-7}      & 7\,\text{px} & 14\,\text{px} & 28\,\text{px} & 56\,\text{px} & 112\,\text{px} & mean \\
\midrule
\multicolumn{1}{l}{ResNet50 (ArcFace~\cite{deng2019arcface})} & \cl 51.08 & 57.74 & \cl 89.43 & 94.59 & \cl 95.01 & 77.57 \\
\multicolumn{1}{l}{\hspace{2mm}+ $\lossOct$ fine-tuning} & \cl 78.20 & 88.31 & \cl 92.97 & 94.56 & \cl 94.65 & 89.74 \\
      & \cl \smaller{(\clg{+27.12})} & \smaller{(\clg{+30.57})} & \cl \smaller{(\clg{+3.54})} & \smaller{(\clr{-0.03})} & \cl \smaller{(\clr{-0.36})} & \smaller{(\clg{+12.17})} \\
\bottomrule
\end{tabular}%
}%
  \label{tab:same-res-comp}%
\end{table}%

In conclusion, these improvements testify to a further contribution toward universal, resolution-independent face recognition systems.

% --------------------------------------------------------------------------------------------------------
\subsubsection{Comparison with other SOTA Approaches}

After demonstrating that our octuplet loss \lossOct improves the robustness of various face recognition models in cross-resolution scenarios, we compare \lossOct with state-of-the-art cross-resolution methods. For this purpose, we evaluated our two best-performing approaches (FaceTransformer~\cite{zhong2021facetransformer} and MagFace~\cite{meng2021magface} with \lossOct) on LFW~\cite{huang2014lfw} with particular resolutions to match the evaluation conditions of other approaches and enable a direct comparison. The results are reported in~\cref{tab:sota-comp-2} and show that our approaches outperform all other methods except for $r=8$\,px image resolution, where Lai and Lam~\cite{lai2021deep} achieved a higher accuracy. However, a notable drawback of their approach is the weak performance for high-resolution images. We must interpret the results of Ge \etal~\cite{ge2018low} carefully as their approach is based on a teacher model that performs worse on high-resolution images (only $97.15\%$) and they report numbers of specific models for each image resolution. Moreover, the training resolution is inconsistent across the compared methods and can lead to slight deviations. Concluding, these results provide a reasonable classification of our approach as the state-of-the-art and underline its advantages.

\begin{table}[h]
  \centering
  \caption{Cross-resolution face verification accuracy [$\%$], evaluated on LFW for different image resolutions. The best accuracy per resolution is marked in bold.}
  \resizebox{0.6\linewidth}{!}{% Table generated by Excel2LaTeX from sheet 'Tables'
\begin{tabular}{lccccc}
\toprule
Model & $8\,$px & $12\,$px & $16\,$px & $32\,$px & high resolution \\
\midrule
Lai and Lam~\cite{lai2021deep} & \cl \textbf{94.8} & 97.6 & \cl 98.2 & $-$   & \cl 99.1 \smaller{($128\,$px)} \\
Sun et al.~\cite{sun2020classifier} & \cl 90.0 & 94.9 & \cl 97.2 & $-$   & \cl 99.1 \smaller{($112$$\times$$96\,$px)} \\
DCR~\cite{lu2018deep} & \cl 93.6 & 95.3 & \cl 96.6 & $-$   & \cl 98.7 \smaller{($112$$\times$$96\,$px)} \\
TCN~\cite{zha2019tcn} & \cl 90.5 & 94.7 & \cl 97.2 & $-$   & \cl 98.8 \smaller{($112$$\times$$96\,$px)} \\
Ge et al.~\cite{ge2018low} & \cl $-$ & $-$   & \cl 85.87 & 89.72 & \cl 97.15 \smaller{($224\,$px)} \\
\midrule
ResNet50 (ArcFace~\cite{deng2019arcface}) & \cl  &       & \cl  &       & \cl  \\
\hspace{2mm}+ $\lossOct$ fine-tuning & \cl 90.38 & 96.88 & \cl 98.28 & 99.48 & \cl 99.55 \smaller{($112\,$px)} \\
FaceTransformer~\cite{zhong2021facetransformer} & \cl  &       & \cl  &       & \cl  \\
\hspace{2mm}+ $\lossOct$ fine-tuning & \cl 94.02 & \textbf{98.17} & \cl \textbf{99.08} & \textbf{99.57} & \cl 99.63 \smaller{($112\,$px)} \\
\bottomrule
\end{tabular}%
}%
  \label{tab:sota-comp-2}%
\end{table}%

In addition, we compare our octuplet loss \lossOct with the approach of Terh\"{o}rst \etal~\cite{terhorst2021qmagface}. While they perform worse on XQLFW ($83.95\%$), they report a slightly higher accuracy on LFW and much better results on AgeDB and CFP-FP. However, they aim at general quality-robust face recognition encompassing resolution, age, and pose. In contrast, we focus exclusively on the images' resolution; hence, this is not a fair comparison and should be considered with caution.

% ========================================================================================================
\subsection{Analysis and Characteristics}
\label{subsec:analysis}

For the analysis, we are using a re-implementation of the popular ArcFace~\cite{deng2019arcface} approach, pre-trained on MS1M-V2~\cite{guo2016ms, deng2019arcface}. It consists of a ResNet50~\cite{he2016resnet} backbone network followed by a $512$-dimensional fully connected layer, which acts as a bottleneck layer during pre-training and provides the facial features $\vec{f}$ for the octuplet loss $\lossOct$. We denote this in the following as our baseline network.

\begin{figure}[b]
    \centering
    \includegraphics[width=0.5\linewidth]{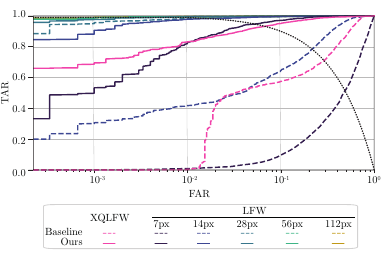}
    \caption{Cross-resolution receiver operating characteristics (ROC) curve comparison of the baseline model (dashed) with our proposed octuplet loss $\lossOct$ fine-tuning (solid) on XQLFW, and LFW for selected image resolutions. The equal error rate (EER) is indicated by a dotted line.}
    \label{fig:roc_curve}
\end{figure}

We provide the receiver operating characteristics curve in~\cref{fig:roc_curve} to obtain deeper insights into the fine-tuning effect on our baseline network on XQLFW~\cite{knoche2021xqlfw} and the LFW~\cite{huang2014lfw} database at different scales. Primarily at very low false acceptance rates (FAR), the performance gain after the fine-tuning is tremendous. While the baseline model fails for challenging situations, the fine-tuned version achieves superior results. On the XQLFW dataset, our approach increases the true acceptance rate (TAR) for very low FARs from $0\%$ to over $65\%$. This improvement is similar to the behavior on the LFW dataset at $7$\,px. The effect vanishes the higher the resolution until, at $112$\,px, the rates remain nearly equal. Overall, this improvement shows the benefit of our method, especially in security applications, \eg, manhunts via surveillance cameras. 

\begin{figure}[h]
    \centering
    \includegraphics[width=0.6\linewidth]{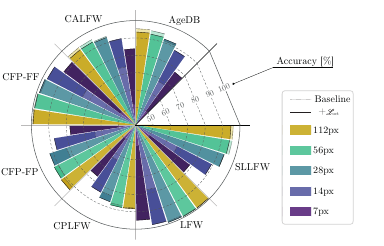}
    \caption{Cross-resolution face verification accuracy comparison of the baseline model and our proposed octuplet loss $\lossOct$ fine-tuning on several datasets (see~\cref{subsec:datasets}) for different image resolutions. An improvement is highlighted with darker colors, whereas a deterioration is indicated with lighter colors.}
    \label{fig:dataset-comp}
\end{figure}

Moreover, we investigate the deviation in the accuracy change between several datasets. In~\cref{fig:dataset-comp}, we fanned out the increase for several datasets and different image resolutions. We observe a significant performance reduction of the baseline model on challenging datasets that focus on age, pose, person similarity, or low image quality. This indicates that the image resolution is even more critical in combination with other adverse conditions. Our proposed octuplet loss fine-tuning strategy accomplishes the best accuracy for LFW~\cite{huang2014lfw} and CFP-FF~\cite{sengupta2016cfp-fp} with over $90\%$ at all resolutions, which are the easiest benchmarks. In contrast, CPLFW~\cite{zheng2018cplfw} seems to be the most challenging dataset, with a performance below $90\%$ at all scales. The chart also reveals that for large pose variations datasets such as CPLFW~\cite{zheng2018cplfw} and CFP-FP~\cite{sengupta2016cfp-fp}, there is still a moderate increase of accuracy at $28\,$px image resolution, whereas the boost at that scale is marginal for all other datasets. Only on data with a large age gap (AgeDB~\cite{moschoglou2017agedb} and CALFW~\cite{zheng2017calfw}) and similar faces (SLLFW~\cite{deng2017sllfw}), our approach marginally reduces the accuracy on $56\,$px and $112\,$px image resolution. 

In conclusion, this analysis uncovers that our proposed approach is not limited to relatively simple datasets like LFW~\cite{huang2014lfw} but also offers immense benefits to more challenging datasets, which involve, \eg, large age gaps or head pose variances.

% ========================================================================================================
\subsection{Ablation Studies}
\label{subsec:ablation}
We conduct multiple ablation studies to understand the influence of our loss terms, distance metric, feature normalization, margins, and batch size. Firstly, we study each single triplet loss term's contribution and then investigate the distance metric's influence, followed by clarifying the impact of normalizing features. Since the margin is crucial in our proposed octuplet loss, we empirically search for the optimal value for fine-tuning the baseline network. This study, combined with the effect of the batch size, is finally presented in this section. As in \cref{subsec:analysis}, we use the re-implementation of ArcFace~\cite{deng2019arcface} with a ResNet50~\cite{he2016resnet} as the pre-trained network for all ablation studies. 

\begin{table*}[t]
  \centering
  \caption{Two ablation studies: Influence on the cross-resolution face verification accuracy $[\%]$ from each triplet loss term (upper part) and the influence of the distance metric and normalization of features (lower part) using our proposed octuplet loss $\lossOct$ fine-tuning evaluated on several datasets (see~\cref{subsec:datasets}) and different image resolutions. The best performance within each study is marked in bold.}
  \resizebox{\linewidth}{!}{% Table generated by Excel2LaTeX from sheet 'Tables'
\begin{tabular}{rccccccccccc}
\toprule
\multicolumn{4}{c}{$\lossTri(\cdot)$} & \multirow{2}[4]{*}{LFW} & \multicolumn{6}{c}{$\varnothing$(LFW, CALFW, CPLFW, SLLFW, CFP-FF, CFP-FP, AgeDB)} & \multirow{2}[4]{*}{XQLFW} \\
\cmidrule{1-4}\cmidrule{6-11}\multicolumn{1}{l}{$\Th$} & $\Thl$ & $\Tlh$ & $\Tl$ &       & 7\,\text{px} & 14\,\text{px} & 28\,\text{px} & 56\,\text{px} & 112\,\text{px} & mean  &  \\
\midrule
\multicolumn{1}{l}{$\checkmark$} & $\checkmark$ & $\checkmark$ & $\checkmark$ & \cl 99.55 & 83.07 & \cl \textbf{90.72} & 93.65 & \cl 94.46 & 94.65 & \cl \textbf{91.31} & 93.27 \\
\midrule
\midrule
\multicolumn{1}{l}{$\checkmark$} &       &       &       & \cl 99.42 \smaller{(\clr{-0.13})} & 61.17 \smaller{(\clr{-21.90})} & \cl 78.77 \smaller{(\clr{-11.95})} & 92.21 \smaller{(\clr{-1.44})} & \cl 94.77 \smaller{(\clg{+0.31})} & 95.11 \smaller{(\clg{+0.46})} & \cl 84.40 \smaller{(\clr{-6.91})} & 80.98 \smaller{(\clr{-12.29})} \\
      & $\checkmark$ &       &       & \cl 99.52 \smaller{(\clr{-0.03})} & 79.93 \smaller{(\clr{-3.14})} & \cl 89.39 \smaller{(\clr{-1.33})} & 93.57 \smaller{(\clr{-0.08})} & \cl 94.57 \smaller{(\clg{+0.11})} & 94.64 \smaller{(\clr{-0.01})} & \cl 90.42 \smaller{(\clr{-0.89})} & 92.15 \smaller{(\clr{-1.12})} \\
      &       & $\checkmark$ &       & \cl 99.48 \smaller{(\clr{-0.07})} & 80.10 \smaller{(\clr{-2.97})} & \cl 89.48 \smaller{(\clr{-1.24})} & 93.53 \smaller{(\clr{-0.12})} & \cl 94.56 \smaller{(\clg{+0.10})} & 94.62 \smaller{(\clr{-0.03})} & \cl 90.46 \smaller{(\clr{-0.85})} & 92.32 \smaller{(\clr{-0.95})} \\
      &       &       & $\checkmark$ & \cl 98.95 \smaller{(\clr{-0.60})} & 82.49 \smaller{(\clr{-0.58})} & \cl 88.80 \smaller{(\clr{-1.92})} & 91.40 \smaller{(\clr{-2.25})} & \cl 92.21 \smaller{(\clr{-2.25})} & 92.06 \smaller{(\clr{-2.59})} & \cl 89.39 \smaller{(\clr{-1.92})} & 93.20 \smaller{(\clr{-0.07})} \\
\midrule
\multicolumn{1}{l}{$\checkmark$} & $\checkmark$ &       &       & \cl 99.57 \smaller{(\clg{+0.02})} & 79.60 \smaller{(\clr{-3.47})} & \cl 89.50 \smaller{(\clr{-1.22})} & 93.93 \smaller{(\clg{+0.28})} & \cl \textbf{95.19} \smaller{(\clg{+0.73)}} & \textbf{95.18} \smaller{(\clg{+0.53)}} & \cl 90.68 \smaller{(\clr{-0.63})} & 91.93 \smaller{(\clr{-1.34})} \\
\multicolumn{1}{l}{$\checkmark$} &       & $\checkmark$ &       & \cl \textbf{99.58} \smaller{(\clg{+0.03)}} & 79.42 \smaller{(\clr{-3.65})} & \cl 89.75 \smaller{(\clr{-0.97})} & \textbf{93.97} \smaller{(\clg{+0.32)}} & \cl 95.11 \smaller{(\clg{+0.65})} & 95.14 \smaller{(\clg{+0.49})} & \cl 90.68 \smaller{(\clr{-0.63})} & 92.18 \smaller{(\clr{-1.09})} \\
\multicolumn{1}{l}{$\checkmark$} &       &       & $\checkmark$ & \cl 99.57 \smaller{(\clg{+0.02})} & 81.66 \smaller{(\clr{-1.41})} & \cl 90.05 \smaller{(\clr{-0.67})} & 93.66 \smaller{(\clg{+0.01})} & \cl 94.65 \smaller{(\clg{+0.19})} & 94.84 \smaller{(\clg{+0.19})} & \cl 90.97 \smaller{(\clr{-0.34})} & 92.77 \smaller{(\clr{-0.50})} \\
      & $\checkmark$ & $\checkmark$ &       & \cl 99.45 \smaller{(\clr{-0.10})} & 80.51 \smaller{(\clr{-2.56})} & \cl 89.90 \smaller{(\clr{-0.82})} & 93.52 \smaller{(\clr{-0.13})} & \cl 94.48 \smaller{(\clg{+0.02})} & 94.54 \smaller{(\clr{-0.11})} & \cl 90.59 \smaller{(\clr{-0.72})} & 92.55 \smaller{(\clr{-0.72})} \\
\midrule
\multicolumn{1}{l}{$\checkmark$} & $\checkmark$ & $\checkmark$ &       & \cl \textbf{99.58} \smaller{(\clg{+0.03)}} & 80.60 \smaller{(\clr{-2.47})} & \cl 90.11 \smaller{(\clr{-0.61})} & 93.94 \smaller{(\clg{+0.29})} & \cl 94.97 \smaller{(\clg{+0.51})} & 94.94 \smaller{(\clg{+0.29})} & \cl 90.91 \smaller{(\clr{-0.40})} & 92.55 \smaller{(\clr{-0.72})} \\
\multicolumn{1}{l}{$\checkmark$} & $\checkmark$ &       & $\checkmark$ & \cl 99.50 \smaller{(\clr{-0.05})} & 82.92 \smaller{(\clr{-0.15})} & \cl \textbf{90.72} \smaller{(0)} & 93.70 \smaller{(\clg{+0.05})} & \cl 94.52 \smaller{(\clg{+0.06})} & 94.60 \smaller{(\clr{-0.05})} & \cl 91.29 \smaller{(\clr{-0.02})} & 93.30 \smaller{(\clg{+0.03})} \\
\multicolumn{1}{l}{$\checkmark$} &       & $\checkmark$ & $\checkmark$ & \cl 99.37 \smaller{(\clr{-0.18})} & 82.49 \smaller{(\clr{-0.58})} & \cl 90.35 \smaller{(\clr{-0.37})} & 93.58 \smaller{(\clr{-0.07})} & \cl 94.47 \smaller{(\clg{+0.01})} & 94.56 \smaller{(\clr{-0.09})} & \cl 91.09 \smaller{(\clr{-0.22})} & \textbf{93.48} \smaller{(\clg{+0.21)}} \\
      & $\checkmark$ & $\checkmark$ & $\checkmark$ & \cl 99.20 \smaller{(\clr{-0.35})} & \textbf{83.09} \smaller{(\clg{+0.02)} & \cl 90.62 \smaller{(\clr{-0.10})} & 93.18 \smaller{(\clr{-0.47})} & \cl 93.93 \smaller{(\clr{-0.53})} & 93.95 \smaller{(\clr{-0.70})} & \cl 90.96 \smaller{(\clr{-0.35})} & 93.42 \smaller{(\clg{+0.15})} \\
\midrule
\midrule
      &       &       &       &       &       &       &       &       &       &       &  \\
\multicolumn{1}{l}{$\deuc$} & $\deucsqa$ & $\dcos$ & $\lVert \vec{f} \rVert_{2}$ &       &       &       &       &       &       &       &  \\
\midrule
\multicolumn{1}{l}{$\checkmark$} &       &       &       & \cl 99.55 & \textbf{83.07} & \cl \textbf{90.72} & 93.65 & \cl 94.46 & 94.65 & \cl \textbf{91.31} & 93.27 \\
\midrule
\multicolumn{1}{l}{$\checkmark$} &       &       & $\checkmark$ & \cl 99.60 \smaller{(\clg{+0.05})} & 79.53 \smaller{(\clr{-3.54})} & \cl 88.64 \smaller{(\clr{-2.08})} & 93.59 \smaller{(\clr{-0.06})} & \cl 94.80 \smaller{(\clg{+0.34})} & 94.95 \smaller{(\clg{+0.30})} & \cl 90.30 \smaller{(\clr{-1.01})} & 93.27 \smaller{(0})} \\
      & $\checkmark$ &       & $\checkmark$ & \cl \textbf{99.63} \smaller{(\clg{+0.08)}} & 80.08 \smaller{(\clr{-2.99})} & \cl 89.63 \smaller{(\clr{-1.09})} & \textbf{93.83} \smaller{(\clg{+0.18)}} & \cl \textbf{95.11} \smaller{(\clg{+0.65)}} & \textbf{95.24} \smaller{(\clg{+0.59)}} & \cl 90.78 \smaller{(\clr{-0.53})} & \textbf{93.58} \smaller{(\clg{+0.31)}} \\
      & $\checkmark$ &       &       & \cl 99.53 \smaller{(\clr{-0.02})} & 81.36 \smaller{(\clr{-1.71})} & \cl 89.75 \smaller{(\clr{-0.97})} & 93.55 \smaller{(\clr{-0.10})} & \cl 94.61 \smaller{(\clg{+0.15})} & 94.71 \smaller{(\clg{+0.06})} & \cl 90.80 \smaller{(\clr{-0.51})} & 93.23 \smaller{(\clr{-0.04})} \\
      &       & $\checkmark$ & $\checkmark$ & \cl 99.58 \smaller{(\clg{+0.03})} & 79.57 \smaller{(\clr{-3.50})} & \cl 88.85 \smaller{(\clr{-1.87})} & 93.70 \smaller{(\clg{+0.05})} & \cl 94.92 \smaller{(\clg{+0.46})} & 95.03 \smaller{(\clg{+0.38})} & \cl 90.41 \smaller{(\clr{-0.90})} & 93.27 \smaller{(0)} \\
\bottomrule
\end{tabular}%
}%
  \label{tab:loss-dist-comp}%
\end{table*}%

% --------------------------------------------------------------------------------------------------------
\subsubsection{Loss Terms}
Our proposed octuplet loss consists of four different triplet loss functions (cf. \cref{eq:Octupletloss}), and each term affects the overall performance. Hence, we conduct experiments to obtain the contribution of each term. In this study, we use the Euclidean distance and no feature normalization. As depicted in the upper part of~\cref{tab:loss-dist-comp}, we start from the best mean accuracy across all datasets and image resolutions, which is obtained by including all triplet loss terms. 

As expected, utilizing only $\lossTri(\Th)$ leads to the worst results on XQLFW~\cite{knoche2021xqlfw} and images with resolutions ($7\,$-$28\,$px), but interestingly, it does not improve accuracy on the high-resolution dataset LFW~\cite{huang2014lfw}, whereas it does improve the accuracy on average across all datasets. We suspect that: 1) The performance is already saturating for LFW, and 2) there might be a few lower-quality images in the LFW dataset, although we expect them to be exclusively in high resolution. However, this term is essential to constrain the network and not focus entirely on low resolution. In contrast, utilizing only $\lossTri(\Tl)$ significantly improves the performance on low-resolution images. Nevertheless, it drastically reduces the verification accuracy on high-quality images and thus is not considered preferable. Considering only $\lossTri(\Thl)$, $\lossTri(\Tlh)$, or the inclusion of both terms, leads to a moderate increase of robustness to image resolution but also comes with the trade-off of reducing the accuracy on high-resolution images. A similar effect occurs for the combination of $\lossTri(\Th)$ and $\lossTri(\Tl)$. Interestingly, the $\lossTri(\Th)+\lossTri(\Thl)$ or $\lossTri(\Th)+\lossTri(\Tlh)$ configuration yields the best performance on intermediate image resolutions ($28\,$px and $56\,$px). Furthermore, experiments with three \lossTri terms reveal that removing $\lossTri(\Th)$ or $\lossTri(\Tl)$ leads to a marginal decline in performance. 

This breakdown of the individual loss terms shows that each term contributes to the overall performance. However, the benefit of including both $\lossTri(\Th)$ and $\lossTri(\Tl)$ instead of simply one of them is only minor since they both connect high-resolution images with low-resolution images (\cf~\cref{fig:1-loss-vis}).

% --------------------------------------------------------------------------------------------------------
\subsubsection{Distance Metric and Feature Normalization}
As described in~\cref{subsec:settings}, our proposed approach follows the work of Hermans \etal~\cite{hermans2017defense} and uses the Euclidean distance metric without feature normalization. However, as proposed in other works~\cite{schroff2015facenet, boutros2022self, parkhi2015deep, feng2020triplet}, we experimented with the squared Euclidean distance. Additionally, we conduct experiments with the cosine distance metric. Since those configurations consequently affect the magnitude of the margin, we empirically determine the best margins for each configuration. 

\Cref{tab:sota-comp-2} illustrates the face verification accuracies and points out that the Euclidean distance is best for low-resolution images. In contrast, for intermediate and high-resolution images (from $28\,$px up to $112\,$px), the Euclidean squared distance and feature normalization leads to the best results. The improvement for this configuration is also evident in the XQLFW~\cite{knoche2021xqlfw} dataset and might be preferred for real-world applications. Due to the utilization of the cosine distance in our evaluation protocols, one would expect that utilizing this metric in our octuplet loss fine-tuning strategy leads to the best results. However, this is not true, as seen in the bottom row of \cref{tab:sota-comp-2}. We can achieve similar performance on LFW~\cite{huang2014lfw} and XQLFW, but for lower image resolutions, fine-tuning with cosine distance leads to a smaller improvement.

% --------------------------------------------------------------------------------------------------------
\subsubsection{Margin and Batch Size}

\begin{figure}[h]
    \centering
    \includegraphics[width=0.7\linewidth]{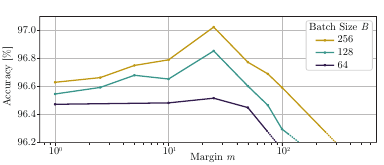}
    \caption{Cross-resolution face verification accuracy $[\%]$ with our proposed octuplet loss $\lossOct$ comparison of different values for margin $m$ and batch-size, evaluated on our validation split of MS1M-V2. Note that we report mean accuracy for images resolutions $r \in \{7, 14, 28, 56, 112\}$.}
    \label{fig:margin-batchsize}
\end{figure}

To conclude our ablation studies, we report the face verification accuracy after fine-tuning our baseline network with different margins $m$ and batch sizes $B$. Keeping in mind that our baseline network achieves $84.90\%$ accuracy, in~\cref{fig:margin-batchsize}, it is obvious that the improvement is more prominent for a larger number of samples within each batch. This effect is unsurprising since a larger batch size increases the probability of the hard sample mining algorithm finding even more challenging samples. Batches containing less than $64$ samples lead to even worse accuracy, and hence, they are not further investigated in this work. Due to hardware limitations, we were unable to conduct experiments with larger batch sizes. Nevertheless, we expect this trend to continue until our hard sample mining algorithm only selects outliers, \ie, incorrectly labeled images. 

In addition, we evaluated the face verification accuracy for margin values between $1$ and $500$. A value of $m = 25$ leads to peak performance. This maximum is also consistent for different batch sizes and indicates that the margin is independent of the batch size. 

\section{Conclusion}
\label{sec:conclusion}
This work conducts further research on low-/cross-resolution face recognition and proposes a novel fine-tuning strategy with an octuplet loss function for existing models to boost their robustness against varying image resolutions. Our contribution involves a combination of four triplet loss terms applied simultaneously to high- and low-resolution images. This interaction exploits not only the relationship between different resolutions of the same image but also between different images of the same identity. The most significant advantage compared to other approaches is that this method can be built on top of existing approaches instead of a costly re-training. 

We demonstrated the effectiveness of our fine-tuning strategy with several state-of-the-art face recognition approaches and observed a vast increase of robustness against image resolution without any significant trade-off on high-resolution images. Our approach performs best on the recently published cross-quality labeled faces in the wild dataset achieving $95.12\%$ accuracy. Additionally, we exhaustively analyzed the improvements on several popular datasets and concluded that our method is universally applicable. Moreover, our ablation study revealed that all four triplet loss terms are needed to perform superior. We discovered that the distance metric and feature normalization plays a less important role as long as the margin for the triplet loss terms is chosen correctly.

Our future work will focus on reducing the amount of data needed for the octuplet loss fine-tuning strategy to further reduce the training time. In other words, we want to follow up on the hard sample mining strategy. An intelligent distillation process of the training set, \ie, keeping only the most relevant images, could potentially achieve even faster convergence. In masked face recognition, we witness many analogies to cross-resolution face recognition, so it would be interesting to explore if our octuplet loss concept could be beneficial there. 

We believe that our contribution can help the community build more robust face recognition systems in the future. Code and details are released under the MIT license.

%%%%%%%%%  REFERENCES
{\small
\bibliographystyle{IEEEbib}
\bibliography{main}
}

\end{document}